\pdfoutput=1

\documentclass[11pt]{article}

\usepackage[]{naacl2021}

\usepackage{times}
\usepackage{latexsym}

\usepackage[T1]{fontenc}

\usepackage[utf8]{inputenc}

\usepackage{microtype}
\usepackage{algorithm,algpseudocode,color,graphicx,multirow,paralist,pifont,subfigure,url,xspace}
\usepackage{amsfonts,amsmath}
\usepackage{booktabs}
\usepackage{setspace}
\definecolor{darkred}{rgb}{0.8,0.0,0.0}
\definecolor{darkblue}{rgb}{0.0,0.0,0.5}
\definecolor{darkgreen}{rgb}{0.0,0.5,0.0}

\usepackage{hyperref}
\usepackage{amssymb}
\usepackage{pifont}
\newcommand{\cmark}{\ding{52}}%
\newcommand{\xmark}{\ding{56}}%
\usepackage{balance}
\newcommand*{\Scale}[2][4]{\scalebox{#1}{$#2$}}

\title{Technical Question Answering across Tasks and Domains}
\date{}

\begin{document}

\author{{\bf Wenhao Yu$^{\dag}$, Lingfei Wu$^{\ddag}$, Yu Deng$^{\ddag}$, Qingkai Zeng$^{\dag}$,} \\
{\bf Ruchi Mahindru$^{\ddag}$, Sinem Guven$^{\ddag}$, Meng Jiang$^{\dag}$} \\
$\dag$University of Notre Dame, Notre Dame, IN, USA\\ 
$\ddag$IBM Thomas J. Watson Research Center, Yorktown Heights, NY, USA\\
{\tt $\dag$\{wyu1, qzeng, mjiang2\}@nd.edu} \\
{\tt $\ddag$\{wuli, dengy, rmahindr, sguven\}@us.ibm.com}
}

\maketitle

\begin{abstract}

Building automatic technical support system is an important yet challenge task. Conceptually, to answer a user question on a technical forum, a human expert has to first retrieve relevant documents, and then read them carefully to identify the answer snippet. Despite huge success the researchers have achieved in coping with general domain question answering (QA), much less attentions have been paid for investigating technical QA. Specifically, existing methods suffer from several unique challenges (i) the question and answer rarely overlaps substantially and (ii) very limited data size. In this paper, we propose a novel framework of deep transfer learning to effectively address technical QA across tasks and domains. To this end, we present an adjustable joint learning approach for document retrieval and reading comprehension tasks. Our experiments on the TechQA demonstrates superior performance compared with state-of-the-art methods.
\end{abstract}

\section{Introduction}
\label{sec:introduction}

Recent years have seen a surge of interests in building automatic technical support system, partially due to high cost of training and maintaining human experts and significant difficulty in providing timely responses during the peak season. Huge successes have been achieved in coping with 
open-domain QA tasks
~\cite{chen2020open}, especially with advancement of large pre-training language models~\cite{devlin2019bert}.
Among them, two-stage retrieve-then-read framework is the mainstream way to solve open-domain QA tasks, pioneered by \cite{chen2017reading}: a retriever component finding a document that might contain an answer from a large collection of documents, followed by a reader component finding the answer snippet in a given paragraph or a document. 
Recently, various pre-training language models (e.g., BERT) have dominated the encoder design for solving different open-domain QA tasks~\cite{karpukhin2020dense,xiong2020pretrained}.

\begin{figure}[t]
    \centering
    {\includegraphics[width=0.49\textwidth]{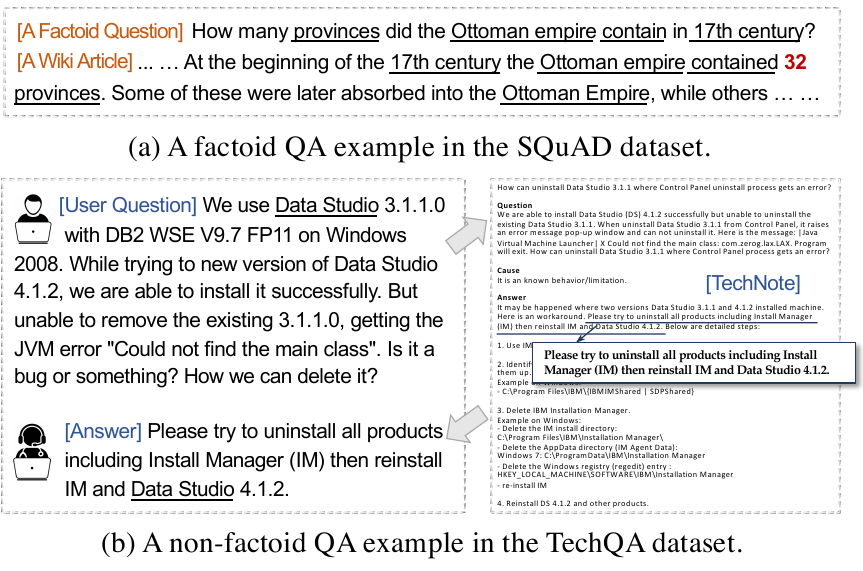}}
    \vspace{-0.3in}
    \caption{Factoid QA is semantic aligned but non-factoid QA has few overlapping words. Semantic similarities between such non-factoid QA is not indicative.}
    \vspace{-0.1in}
    \label{fig:example} 
\end{figure}

Despite the tremendous successes achieved in general QA domain, technical QA have not yet been well investigated due to several unique challenges. 
First, technical QAs are non-factoid. The question and answer can hardly overlap substantially, because the answer typically fills in missing information and actionable solutions to the question such as steps for installing a software package and configuring an application. Different from factoid questions that are typically aligned with a span of text in document~\cite{rajpurkar2016squad,rajpurkar2018know}, semantic similarities between such non-factoid QA pairs could have a large gap as shown in Fig.\ref{fig:example}. 
Therefore, the retrieval module in retrieve-then-read framework might find documents that do not contain correct answers due to the semantic gap in non-factoid QAs~\cite{karpukhin2020dense,lee2019latent,yu2020crossing}.
Second, compared to SQuAD (with more than 100,000 QA pairs), technical domain datasets typically have a much smaller number of labelled QA pairs (e.g., about 1,400 in TechQA), partially due to the prohibitive cost of creating labelled data. In addition, there are limited real user questions and technical support documents, especially for some new tech products and communities. Since the pre-trained language models are mainly trained on general domain corpora, directly fine-tuning pre-trained language models may lead to unsatisfying performance due to the large discrepancy between source tasks (general domains) and target tasks (technical domains)~\cite{chang2020pre,gururangan2020don}.

To address the aforementioned challenges, we propose a novel deep transfer learning framework that explores knowledge \underline{trans}fer across \underline{t}asks and \underline{d}omains (TransTD). TransTD consists of two components: TransT (knowledge transfer across tasks) and TransD (knowledge transfer across domains). 
TransTD jointly learns snippet prediction (reading comprehension) task and matching prediction (document retrieval) task simultaneously, applying it on both general domain QA and target domain QA.

To address the first challenge of non-factoid QAs, TransT leverages a joint learning model that directly ranks all predicted snippets by reading each pair of query and candidate document. It optimizes matching prediction and snippet prediction in parallel. Compared to two-stage retrieve-then-read methods that only read most semantically related documents, TransT considers potential snippets in every candidate document.
When jointly training these two tasks, snippet prediction pays attention to local correspondence and matching prediction helps understand the semantic relationship from a global perspective, allowing the multi-head attentions in BERT-based encoders to jointly attend to information from different representation subspaces at different positions. Besides, the weights of two training objectives can be dynamically learned to pay more attention on the more difficult task when training different data samples.

To address the second challenge of learning with limited data, TransD leverages a deep transfer learning model to transfer knowledge from general domain QAs to technical domain QAs. General domain QA dataset like SQuAD has a much larger data size and a similar task setting (i.e., snippet prediction). Though knowledge is different between two domains, by learning the ability to answer questions in general domains, the model can quickly adapt and learn efficiently when changing into a new domain, reflected in faster convergence and better performance.
Transfer learning helps avoid overfitting on technical QAs with limited size of data. Specifically, our model first applies the multi-task joint learning in general domain QAs (SQuAD), then transfers model parameters to initialize the training in the target domain QAs (TechQA), making knowledge transfer across domains to address data limitation.

We conducted extensive experiments on the TechQA dataset and utilized BERT as basic models. Experiments show that TransTD can provide superior performance than models with no knowledge transfer and other state-of-the-art methods.

\section{Related Work}
\label{sec:related_work}
\paragraph{Open-Domain QA}
Open-domain textual question answering is a task that requires a system to answer factoid questions using a large collection of documents as the information source, without the need of pre-specifying topics or domains~\cite{chen2020open}. Two-stage retriever-reader framework is the mainstream way to solve open-domain QA, pioneered by \cite{chen2017reading}. Recent work has improved this two-stage open-domain QA from different perspectives such as novel pre-training methods~\cite{lee2019latent,guu2020realm}, semantic alignment between question and passage~\cite{lee2019latent,karpukhin2020dense,wu2018word}, cross-attention based BERT retriever~\cite{yang2019end,gardner2019making}, global normalization between multiple passages~\cite{wang2019multi}.



\paragraph{Transfer Learning} 

Transfer learning studies how to transfer knowledge from auxiliary domains to a target domain~\cite{pan2009survey,jiang2015social,yao2019graph}. Recent advances of deep learning technologies with transfer learning has achieved great success in a variety of NLP tasks~\cite{ruder2019transfer}. 
Several research work in this domain greatly enrich the application and technology of transfer learning on question answering from different perspectives~\cite{min2017question,deng2018knowledge,castelli2020techqa,yu2020technical}. 
Although transfer learning has been successfully applied to various QA applications, its applicability to technical QA has yet to be investigated.
In this work, we focus on leveraging transfer learning to enhance QA in tech domain.

\section{Research Problem}
\label{sec:method}

In the technical support domain, suppose we have a set of questions $\mathcal{Q}$ and a large collection of documents $\mathcal{D}$. For each question $Q \in \mathcal{Q}$, we aim at finding a relevant document $D \in \mathcal{D}$ and extracting the snippet answer $S = (D_{start}, D_{end})$ in the document $D$. Note that the answer may not exist, and so, the relevant document may not exist, either. 
All predicted snippets are ranked by a specific span score calculation method, and (usually) the top-1\footnote{Since technical domain RC is extremely difficult, we also evaluate performance on top-5 predictions in our experiments.} answer span is chosen to answer the given question.

\begin{figure*}[t]
    \centering
    {\includegraphics[width=1.0\textwidth]{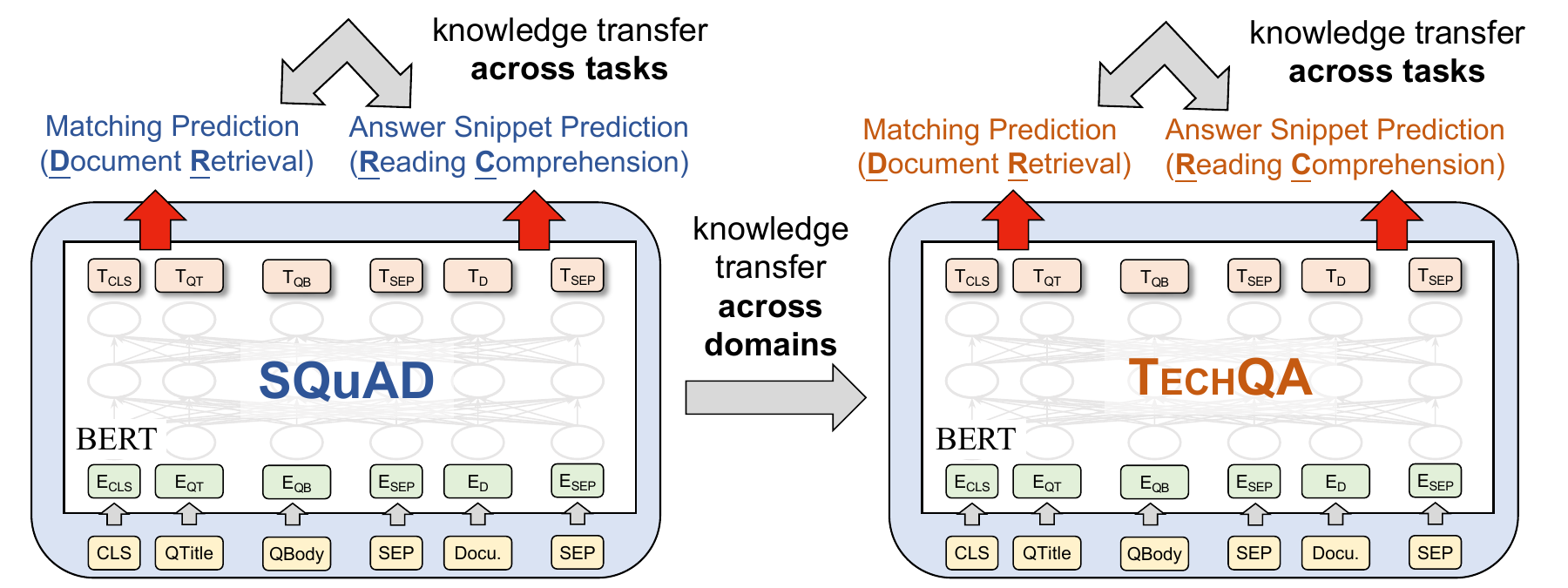}}
    \vspace{-0.2in}
    \caption{Our framework performs knowledge transfer across tasks and domains. It explores the mutual enhancement between the snippet prediction (reading comprehension) and matching prediction (document retrieval), applying multi-task learning to the BERT models on both auxiliary domain (SQuAD) and target domain (TechQA).}
    \label{fig:framework}
\end{figure*}

\section{Proposed Framework}
\label{sec:method}
In this section, we present our proposed framework for technical QA. Given a query, we first obtain \underline{50 Technotes} 
by issuing the query to the search engine Elasticsearch\footnote{Elasticsearch -- https://www.elastic.co/elasticsearch/}.
Instead of using a document retriever based on semantic similarity between the query and each document, our proposed TransTD jointly optimizes snippet prediction and matching prediction in a parallel style. Figure~\ref{fig:framework} illustrates the design of the framework. 
It has a multi-task learning method to transfer knowledge across the snippet prediction (reading comprehension) and matching prediction (document retrieval) tasks. 
This method is further applied to pre-train the model on auxiliary domain QAs\footnote{In our work, auxiliary domain QAs are from general domain QAs, so we use these two words interchangeably. }. Furthermore, the weights of two training objectives are dynamically adjusted by calculating the difference between real answer snippet and predicted snippet.
So, the model can focus on optimizing the more difficult task when training different data samples. 
Lastly, Our model has a novel snippet ranking function that uses snippet prediction to obtain an alignment score and linearly combines it with the matching prediction score.


\subsection{Knowledge Transfer across Tasks}
\label{sec:across_tasks}

We build our model upon BERT \cite{devlin2019bert} to jointly optimize on the RC and DR tasks. Suppose $\Theta$ has the BERT encoder parameters. When we apply domain knowledge transfer, which will be introduced in the following section, we initialize it with the parameters $\Theta^{(aux)}$ trained on the auxiliary domain; when we do not apply the transfer, we initialize it with the original pre-trained BERT parameters. We have two multi-layer perceptron (MLP) classifiers for the two tasks, whose parameters are denoted by $\theta_{RC}$ and $\theta_{DR}$, respectively. Both classifiers are randomly initialized. More specifically, the $RC$ classifier is to predict answer snippets, and the $DR$ classifier is to predict document matching. The joint loss is as follows:
\begin{eqnarray}
    \mathcal{L}^{(aux)} & = & \mathcal{L}_{RC}(\Theta^{(aux)}, \theta^{(aux)}_{RC}) \nonumber \\
    & & + \lambda^{(aux)} \cdot \mathcal{L}_{DR}(\Theta^{(aux)}, \theta^{(aux)}_{DR}),
    \label{eq:loss}
\end{eqnarray}
where $\lambda$ is a hyper-parameter for the weight of the DR task over RC task.

\paragraph{Calculate adjustment factor} 

As shown in Eq.(\ref{eq:loss}), the weights between two training objectives are only adjusted by a pre-determined hyperparameter $\lambda$. However, for different samples in the dataset, the difficulty of learning snippet prediction and matching prediction is different. The weight of two training objectives should be dynamically adjusted so that the model can focus on optimizing the more difficult task when training different data samples. Since non-factoid questions are open-ended questions that often require complex answers that are mostly sentence-level texts, positional relationships between start token and end token in answer snippets have more fluctuations than factoid answers.
Therefore, we take the difference between real answer snippet and predicted snippet to measure the difficulty of snippet prediction.
Intuitively, when the predicted answer snippet is significantly different from the actual answer snippet (much larger or much smaller), it indicates snippet prediction is difficult for the current data sample. So, the model should focus on optimizing the reading comprehension part. On the contrary, the model should focus on optimizing the document retrieval part. Formally, the weight-adjustable joint learning loss function is defined as:
\begin{eqnarray}
    \mathcal{L}^{(aux)} & = & w \cdot \mathcal{L}_{RC}(\Theta^{(aux)}, \theta^{(aux)}_{RC}) \nonumber \\
    & & + \lambda^{(aux)} \cdot \mathcal{L}_{DR}(\Theta^{(aux)}, \theta^{(aux)}_{DR}),
\end{eqnarray}
\begin{eqnarray}
    \Scale[1.1]{w = \exp (\frac{|(D_{end} - D_{start}) - (\hat{D}_{end} - \hat{D}_{start})|}{D_{end} - D_{start}}).}
\end{eqnarray}

\subsection{Knowledge Transfer across Domains}

Besides transferring across tasks, in our framework, we employ knowledge transfer across domains. We identify a dataset from an auxiliary domain (not a technical support domain) for technical question answering like SQuAD. We apply the multi-task learning to the auxiliary domain. The goal is to learn BERT encoder parameters $\Theta^{(aux)}$ and two MLP classifiers $\theta^{(aux)}_{RC}$ and $\theta^{(aux)}_{DR}$:
\begin{eqnarray}
    \mathcal{L}^{(aux)} & = & \mathcal{L}_{RC}(\Theta^{(aux)}, \theta^{(aux)}_{RC}) \nonumber \\
    & & + \lambda^{(aux)} \cdot \mathcal{L}_{DR}(\Theta^{(aux)}, \theta^{(aux)}_{DR}),
\end{eqnarray}
Here the encoder is initialized by the original pre-trained BERT parameters. We will initialize the BERT encoder in the target domain $\Theta$ with $\Theta^{(aux)}$ (used in TransTD-Mean and TransTD-CLS). When $\lambda^{(aux)} = 0$, we apply the single RC task on the auxiliary domain (used in TransTD-single).

\subsection{Framework Components}
\label{sec:components}


\paragraph{Question and Document Encoder} Given a pair of question $Q$ and document $D$, we first build a concatenation by $[[\mathrm{CLS}], Q, [\mathrm{SEP}], D, [\mathrm{SEP}]]$, where $[\mathrm{CLS}]$ stands for a classification token and $[\mathrm{SEP}]$ separates components in the sequence. 
The ${\mathrm{BERT}}_\Theta$ encoder generates contextualized representations of every token $X$ in the input sequence $q$, which is denoted by $\mathrm{BERT}_\Theta (q)[X] \in \mathbb{R}^d$, where $d = 1024$. So we have a matrix of token representations $\mathbf{H} \in \mathbb{R}^{m \times d}$, where $H(k) = \mathrm{BERT}_\Theta (q)[q[k]]$ ($k$ is the index of the token).

\begin{table*}[t]
\centering
\caption{Statistics of TechQA. The test set is not publicly available, only allowing people to submit models for evaluation. The length of \textit{TechNotes} is much bigger than that of question and answer texts.
}
\vspace{-0.1in}
\label{tab:stoa-result}
\setlength{\tabcolsep}{4.8mm}{
\scalebox{0.85}{
\begin{tabular}{l|cccccc}
\toprule
& \#Ques. (answerable/non-ans.) & \#TechNotes & Len-Ques. & Len-Ans. & Len-Notes \\
\midrule
Train & 600 (450 / 150) & 30,000 & 52.1$\pm$31.6 & 48.1$\pm$38.7 & 433.9$\pm$320.6 \\ 
Dev. & 310 (160 / 150) & 15,500 & 53.1$\pm$30.4 & 41.2$\pm$27.7 & 449.1$\pm$351.2  \\
Test & 490 & 24,500 & - & - & - \\
\bottomrule
\end{tabular}}}
\label{tab:stats}
\end{table*}

\begin{table*}[t]
\caption{Ablation study on knowledge transfer across tasks and across domains on TechQA. TransTD transfers knowledge across both tasks and domains, and TransTD$^{+}$ is further improved by the adjustable weight.}
\vspace{-0.1in}
\label{tab:stoa-result}
\centering
\setlength{\tabcolsep}{1.5mm}{
\scalebox{0.85}{%
\begin{tabular}{lcc|cc|ccc|ccc}
\toprule
\multicolumn{2}{c}{\multirow{2}{*}{Methods}} & \multirow{2}{*}{Adjustable} &  \multirow{2}{*}{Source task(s)} & \multirow{2}{*}{Target task(s)} &\multicolumn{3}{c|}{Reading Comprehension} & \multicolumn{3}{c}{Document Retrieval} \\
& & & & & Ma-F1 & HA-F1@1 & HA-F1@5 & MRR & R@1 & R@5 \\ 
\midrule
\multirow{1}{*}{BERT$_{\textbf{DR}}$} & - & \xmark & - & DR & - & - & - & 55.80 & 45.58 & 58.23 \\
\multirow{1}{*}{BERT$_{\textbf{RC}}$} & - & \xmark & - & RC & 52.49 & 24.92 & 37.26 & 51.20 & 48.13 & 56.25  \\
\midrule
\multirow{2}{*}{TransD} & - & \xmark & RC & DR & - & - & - & 60.63 & 58.13 & 64.38 \\
& - & \xmark & RC & RC & 55.31 & 34.69 & 50.52 & 64.60 & 60.63 & 68.23  \\
\multirow{2}{*}{TransT} & CLS & \xmark & - & RC+DR & 53.43 &
26.83 & 38.50 & 51.19 & 46.88 & 56.88 \\
& Mean & \xmark & - & RC+DR & 52.30 & 26.28 & 41.50 & 52.68 & 47.50 & 59.35 \\
\midrule
\multirow{2}{*}{TransTD} & CLS & \xmark & RC+DR & RC+DR & 56.43  &
39.12 & 52.30 &  66.79  & 64.38 & 70.63 \\
& Mean & \xmark & RC+DR & RC+DR & 56.88  & 37.96 & 49.83 & 67.55 & \underline{\textbf{67.50}} & 69.38 \\
\multirow{2}{*}{TransTD$^{\textbf{+}}$} & CLS & \cmark & RC+DR & RC+DR & 56.66 &
38.33 & 50.95 & 67.80  & 65.00 & 72.50 \\
& Mean & \cmark & RC+DR & RC+DR & \underline{\textbf{58.58}} &
\underline{\textbf{40.28}} & \underline{\textbf{52.57}} &  \underline{\textbf{67.98}}  & 66.88 & \underline{\textbf{73.13}} \\
\bottomrule
\end{tabular}}}
\label{tab:ablation}
\end{table*}

\begin{table}[t]
\centering
\caption{TransTD outperforms two-stage retrieve-then-read methods that retrieve document based on semantic alignment. k is the number of retrieved documents.}
\setlength{\tabcolsep}{1mm}{
\scalebox{0.75}{%
\begin{tabular}{cc|ccc}
\toprule
Method & Setting & Ma-F1 & HA-F1@1 & R@1\\ 
\midrule
\multirow{1}{*}{BERTserini~\cite{yang2019end}} & k=1 & 51.34 & 15.23 & 30.00 \\
(with BM25 as retriever) & k=5 & 56.60 & 28.31 & 48.75 \\
\midrule
\multirow{1}{*}{DPR ~\cite{karpukhin2020dense}} & k=1 & 53.22 & 15.57 & 26.25 \\
(w/o pre-trained retriever) & k=5 & 56.47 & 30.40 & 47.50 \\
\midrule
\multirow{1}{*}{DPR ~\cite{karpukhin2020dense}} & k=1 & 54.82 & 19.46 & 30.63 \\
(with pre-trained retriever) & k=5 & 58.56 & 33.03 & 53.13 \\
\midrule
\multirow{2}{*}{TransTD-Mean$^{+}$ (Ours, $\mathrm{S_{\emph{with}}}$)} & \multirow{2}{*}{-} & \multirow{2}{*}{\underline{\textbf{58.58}}} & \multirow{2}{*}{\underline{\textbf{40.28}}} & \multirow{2}{*}{\underline{\textbf{66.88}}} \\ &&&& \\
\bottomrule
\end{tabular}}}
\label{tab:retrieve-then-read}
\end{table}

\paragraph{Reader MLP} This classifier reads the representation matrix $\mathbf{H}$ and computes the score of each token being the start token in the answer snippet $\mathbf{p}_{start} \in \mathbb{R}^{m}$ and the score of each token being the end token $\mathbf{p}_{end} \in \mathbb{R}^{m}$.
\begin{eqnarray}
    \mathbf{p}_{start} = \mathbf{w}_{start} \cdot \mathbf{H}^{\mathrm{T}}, ~\mathbf{p}_{end} = \mathbf{w}_{end} \cdot \mathbf{H}^{\mathrm{T}},
\end{eqnarray}
where $\mathbf{w}_{start}, \mathbf{w}_{end} \in \mathbb{R}^{d}$ are trainable parameters. We have the snippet $S_{RC} = (\hat{D}_{start}, \hat{D}_{end})$ as
\begin{eqnarray}
\hat{D}_{start} & = & {\mathop{\mathrm{argmax}}}_{k \in \{1, \dots, m\}} p_{start}[k], \\
\hat{D}_{end} & = & {\mathop{\mathrm{argmax}}}_{k \in \{1, \dots, m\}} p_{end}[k].
\end{eqnarray}




\paragraph{Matching MLP} Suppose we have the representation of the sequence $q$. It can be denoted by $\mathbf{h} \in \mathbb{R}^d$. The classifier is to predict whether the question $Q$ and document $D$ are aligned, which is a binary variable projected from $\mathbf{h}$:
\begin{equation}
    p_{DR} = \sigma(\mathbf{w}_{DR} \cdot \mathbf{h}), 
\end{equation}
where $\sigma$ is the sigmoid function and $\mathbf{w}_{DR} \in \mathbb{R}^d$ are trainable parameters.
We have two options to produce $\mathbf{h}$ from the input sequence $q$. The first option is to apply mean pooling to the representations of all tokens (used in {TransTD-Mean}):
\begin{equation}
    \mathbf{h} = \textsc{Mean}(\{\mathrm{BERT}_\Theta (q)[X] | X \in q\}).
\end{equation}
The second option is to use the classification token $[\mathrm{CLS}]$ (used in {TransTD-CLS}):
\begin{equation}
    \mathbf{h} = \mathrm{BERT}_\Theta (q)[\mathrm{CLS}].
\end{equation}







\paragraph{Joint Inference} The reading MLP takes question and document pairs and predicts a reading score,
\begin{eqnarray}
    S_{reader} & = & (p_{start}[D_{s}] + p_{end}[D_{e}]) \nonumber \\
    & &  - (p_{start}[0] + p_{end}[0]).
\end{eqnarray}
where $p_{(\cdot)}[0]$ denotes the probability of taking first token of the sequence as the start position or end position of the snippet. The joint ranking score of a ($Q$, $D$) pair is a linear combination of reading score and matching score,
\begin{eqnarray}
    S & = & \alpha \cdot p_{DR} + (1-\alpha) \cdot S_{reader}.
    \label{eq:joint}
\end{eqnarray}

It should be noted that different from previous work that only leverages the first term in reading score, i.e., ${S_{reader} = (p_{start}[D_{s}] + p_{end}[D_{e}])}$~\cite{xiong2020pretrained,qu2020open}, our added second term improved inference performance. This is because during the training time, the span label of a document that does not contain an answer is set to $(0, 0)$, and such negative documents are the majority. Therefore, $(p_{start}[0] + p_{end}[0])$ reflects the probability that $Q$ and $D$ is not aligned. See Table \ref{tab:infer} for experimental comparisons.



\section{Experiments}
\label{sec:experiment}

\subsection{TechQA Dataset}

The TechQA dataset~\cite{castelli2020techqa} contains actual questions posed by users on the IBM DeveloperWorks forums. 
TechQA is designed for machine reading comprehension tasks, 
Each question is associated with a candidate list of 50 \textit{Technotes} obtained by issuing a query on the search engine Elasticsearch\footnote{https://www.elastic.co/elasticsearch/}. A question is answerable if an answer snippet exists in the 50 \textit{Technotes}, or is unanswerable otherwise. 
Data statistics are given in Table \ref{tab:stats}. 
In TechQA, the training set has 600 questions in which 450 questions are answerable; the validation set has 310 questions in which 160 questions are answerable; the test set has 490 questions. The Technotes are usually of greater length than question and answer texts.

\begin{table}[t]
\centering
\caption{Our proposed snippet ranking function can bring additional improvements. Using $(p_{s}[0] + p_{e}[0])$ reflects the degree of misalignment between $Q$ and $D$.}
\setlength{\tabcolsep}{1.5mm}{
\scalebox{0.75}{%
\begin{tabular}{c|ccc}
\toprule
Snippet ranking method & Ma-F1 & HA-F1@1 & R@1 \\ 
\midrule
\multirow{1}{*}{MP-BERT~\cite{wang2019multi}} & \multirow{2}{*}{49.45} & \multirow{2}{*}{24.65} & \multirow{2}{*}{43.75} \\
($\mathrm{S_{MP\text{-}BERT}} = p_{\mathrm{DR}} \cdot p_{\mathrm{s}} \cdot p_{\mathrm{e}}$) &&& \\
\midrule
\multirow{1}{*}{WKLM ~\cite{xiong2020pretrained}} & \multirow{2}{*}{57.82} & \multirow{2}{*}{39.71} & \multirow{2}{*}{66.25} \\
($\mathrm{S_{BERT}} = \alpha \cdot p_{\mathrm{DR}} + p_{\mathrm{s}} + p_{\mathrm{e}}$) &&&  \\
\midrule
\multirow{1}{*}{Ours (w/o document score)} & \multirow{2}{*}{\underline{\textbf{58.58}}} & \multirow{2}{*}{\underline{\textbf{40.28}}} & \multirow{2}{*}{65.00} \\
($\mathrm{S_{\emph{w/o}}} = p_{\mathrm{s}} + p_{\mathrm{e}} - p_{s}[0] - p_{e}[0]$ ) &&&\\
\midrule
\multirow{1}{*}{Ours (with document score)} & \multirow{2}{*}{\underline{\textbf{58.58}}} & \multirow{2}{*}{\underline{\textbf{40.28}}} & \multirow{2}{*}{\underline{\textbf{66.88}}} \\
($\mathrm{S_{\emph{with}}} =  \alpha \cdot p_{\mathrm{DR}} + \mathrm{S_{\emph{w/o}}}$) &&& \\
\bottomrule
\end{tabular}}}
\label{tab:infer}
\end{table}

\begin{figure*}[t]
	\centering
	{\includegraphics[width=0.245\textwidth]{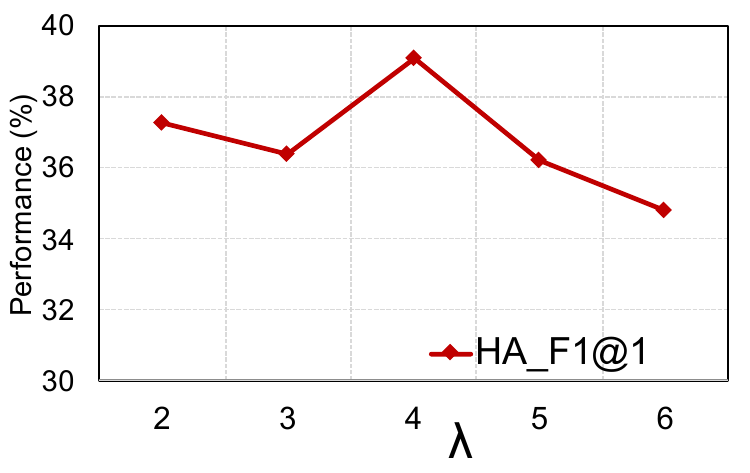}}
	{\includegraphics[width=0.245\textwidth]{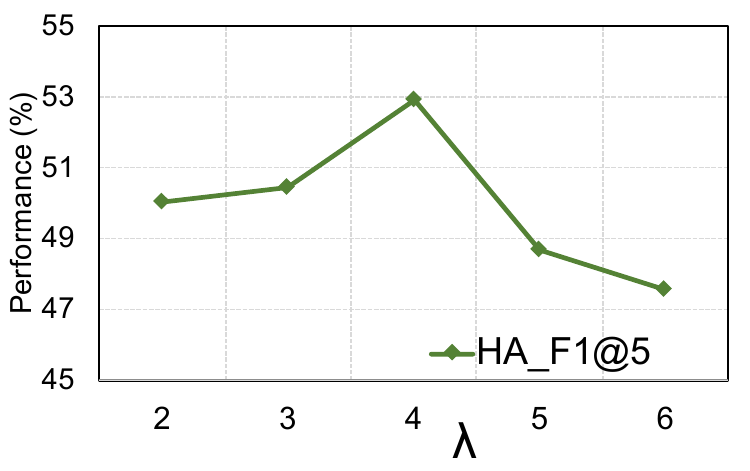}}
	\vline
	{\includegraphics[width=0.245\textwidth]{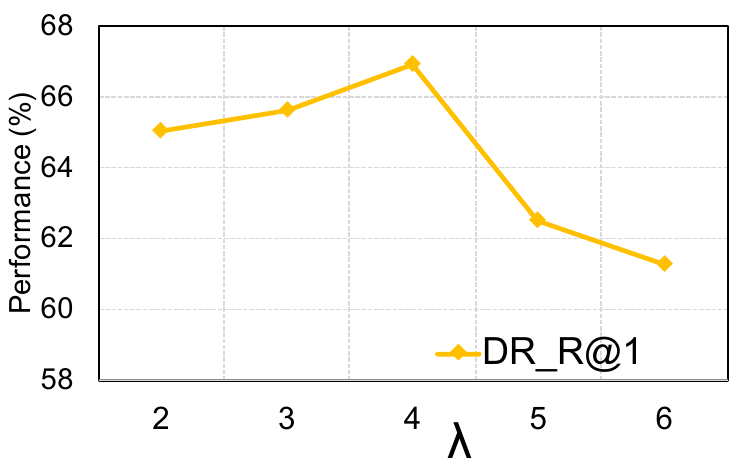}}	
	{\includegraphics[width=0.245\textwidth]{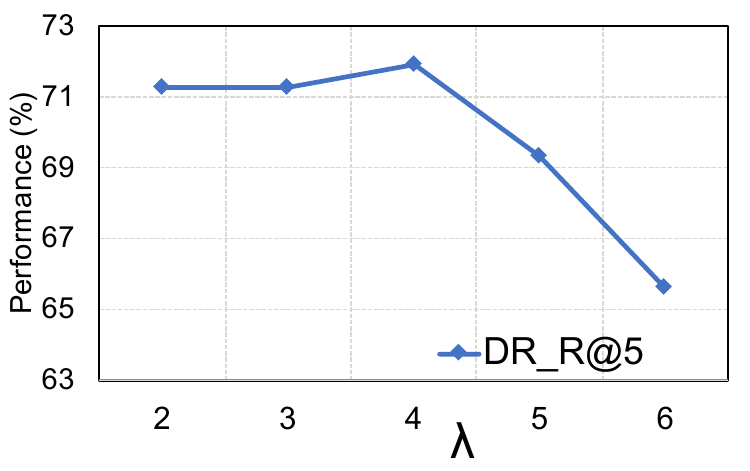}}
	\vspace{-0.25in}
	\caption{$\lambda$ is the weight of the DR task loss over the RC task loss. When $\lambda = 4.0$, TransTD achieves the best performance for both RC (left two) and DR (right two) tasks.}
    \label{fig:labelloss}
\end{figure*}


\begin{figure*}[!t]
	\centering
	{\includegraphics[width=0.245\textwidth]{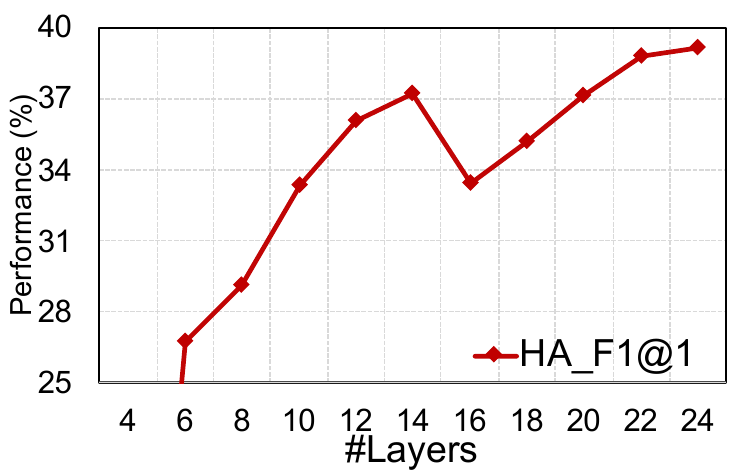}}
	{\includegraphics[width=0.245\textwidth]{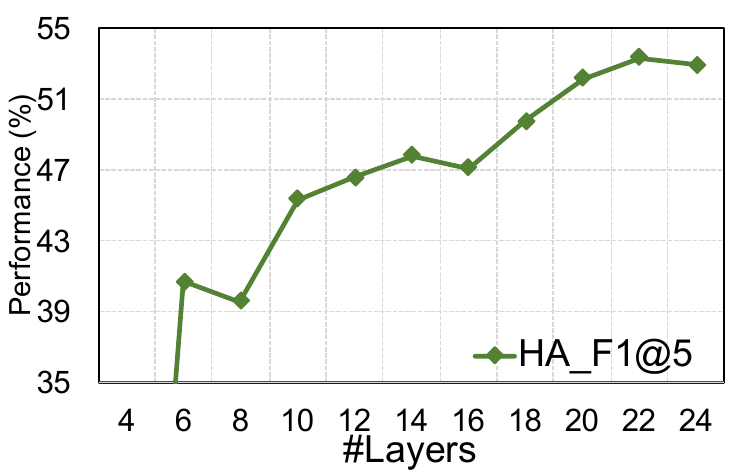}}
	\vline
	{\includegraphics[width=0.245\textwidth]{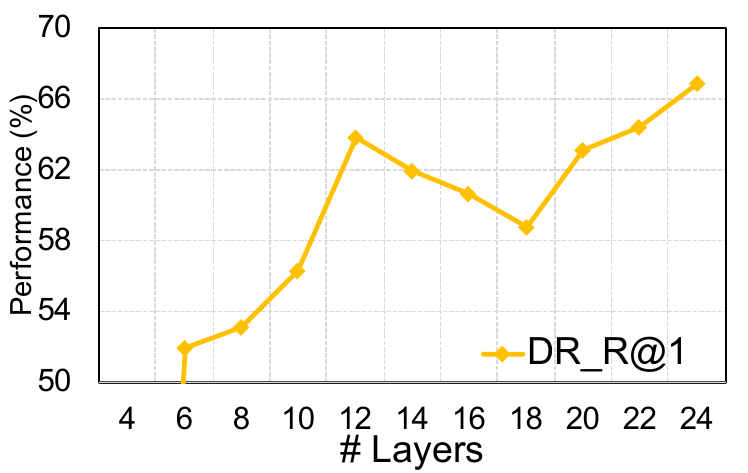}}
	{\includegraphics[width=0.245\textwidth]{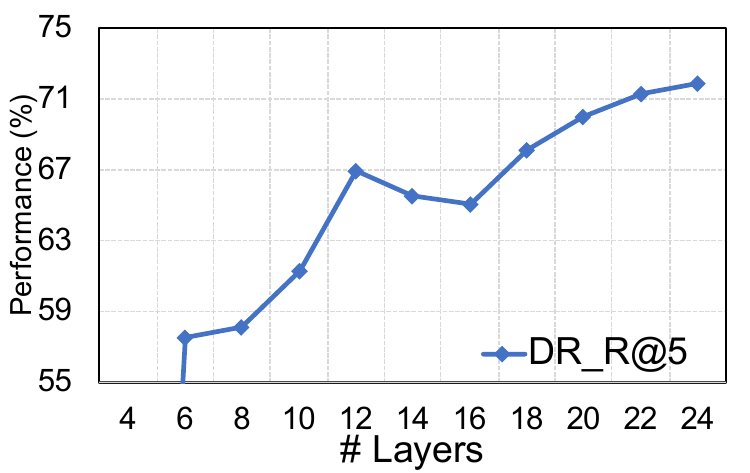}}
	\vspace{-0.25in}
	\caption{The more layers being fine-tuned in the target domain, the better performance we can have. However, it shows the pattern but not always true in the middle of the range.}
	\label{fig:ft_strategy}
\end{figure*}

\begin{figure*}[t]
	\centering
	{\includegraphics[width=0.46\textwidth]{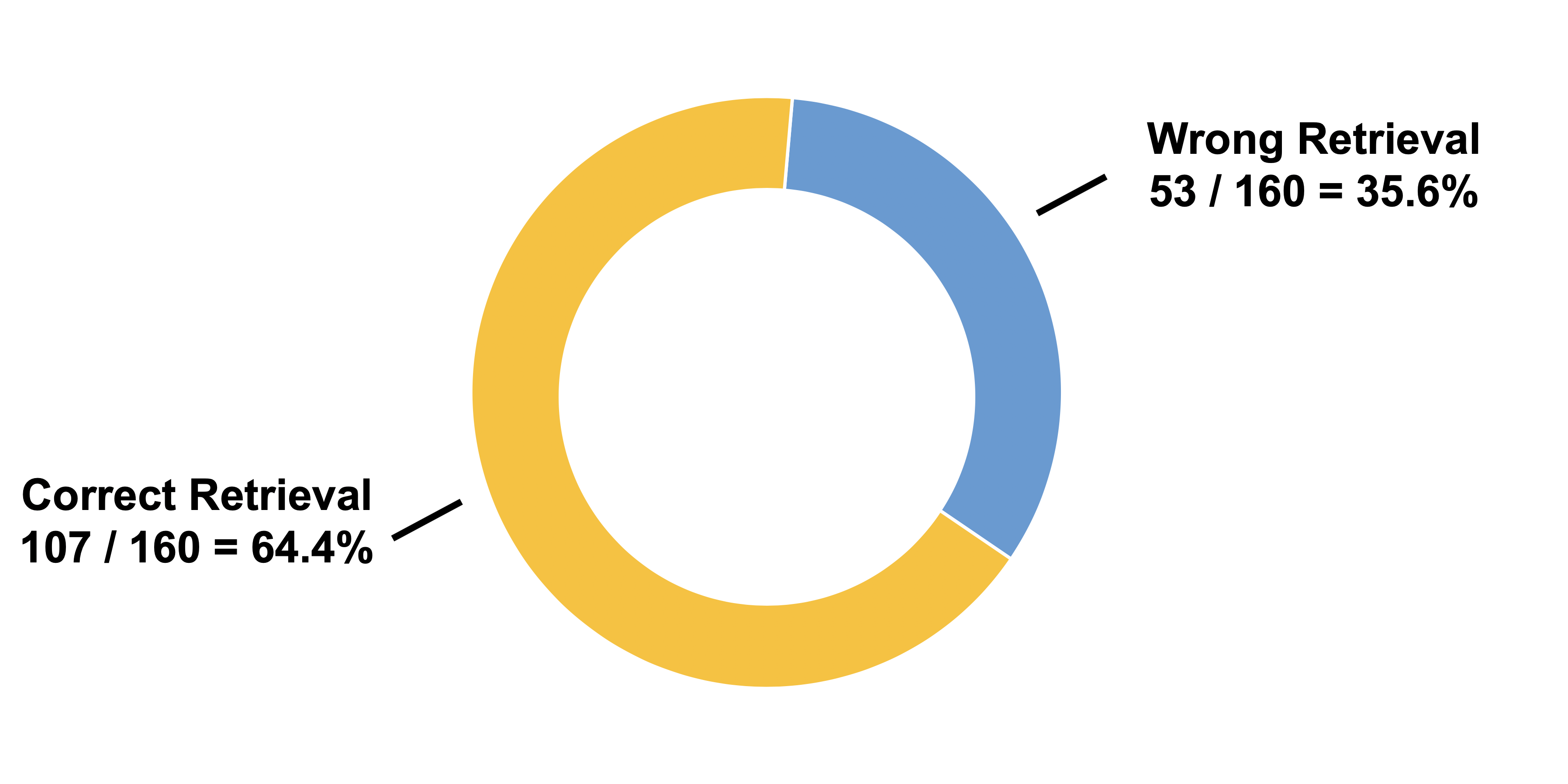}}
	{\includegraphics[width=0.46\textwidth]{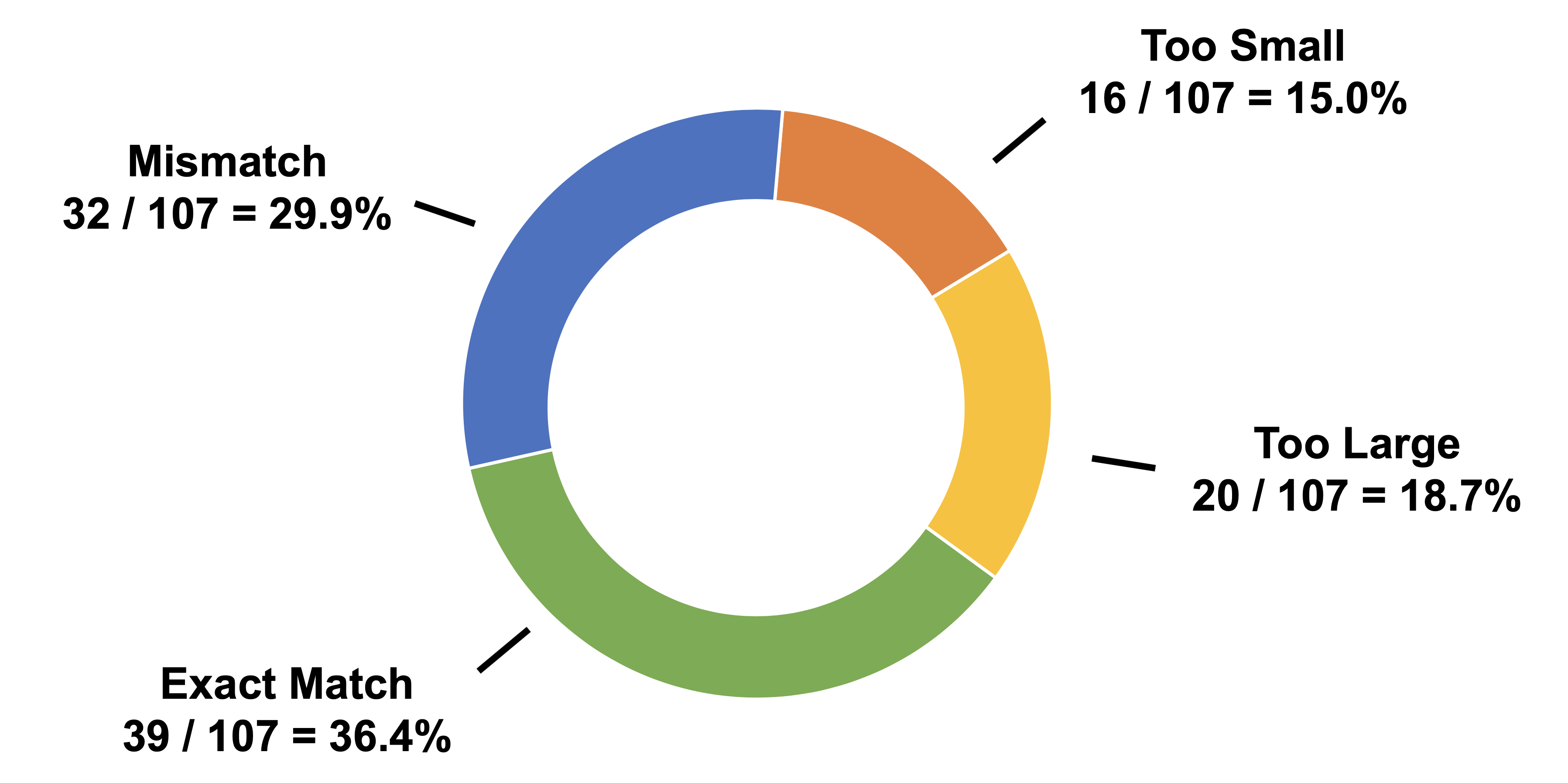}}
	\caption{Error analysis. The left figure represents the proportions between correct and wrong prediction on DR. The right figure represents the proportion of RC results when the retrieval phase already predicts the correct document. (Here, ``too small'' means that if the prediction is $S_{RC} = (D^{(pred)}_{start}, D^{(pred)}_{end})$ and the truth is $S = (D_{start}, D_{end})$, we have $D^{(pred)}_{start} > D_{start}$ and $D^{(pred)}_{end} < D_{end}$; on the contrary, ``too large'' means we have $D^{(pred)}_{start} < D_{start}$ and $D^{(pred)}_{end} > D_{end}$.)}
	\label{fig:error}
\end{figure*}

\subsection{Evaluation methods}

The accuracy of the extracted snippets is evaluated by Ma-F1{\footnote{To avoid confusion between F1 (used on the TechQA leaderboard) and F1@K, we use Ma-F1 instead of F1.}} and HA$\_$F1@$K$. Ma-F1  is the macro average of the F1 scores computed on the first of the $K$ answers provided by the system for each given question:
\begin{equation}
    \mathrm{Ma}\text{-}\mathrm{F1} = \frac{\sum_{i=1}^{K} \mathrm{F1}@K}{K},
\end{equation}
where F1@$K$ computes F1 scores for top-$K$ answer snippets, selects the maximum F1 score, and computes the macro F1 score average over all questions. HA$\_$F1@$K$ calculates macro F1 score average over all answerable questions. Besides, models are evaluated on retrieving and ranking document by mean reciprocal rank (MRR) and recall at $K$ (R@$K$). R@$K$ is the percentage of correct answers in top $K$ out of all the relevant answers. MRR represents the average of the reciprocal ranks of results for a set of queries.

\subsection{Ablation Study}

\noindent \textbf{TranT} transfers knowledge across tasks on the target domain, with multi-tasks of RC and DR.

\noindent \textbf{TranD} transfers knowledge from source domain RC to target domain RC w/o multi-task learning.

\noindent \textbf{TransTD} transfers knowledge across both tasks and domains. TransTD$^{+}$ is further improved by the adjustable weight.

\subsection{Experimental Analysis}

\subsubsection{Knowledge transfer across domains} In Table \ref{tab:ablation}, the model first fine tuning on the source domain QA (SQuAD) then further fine tuning on the target domain QA (TechQA) makes superior performance than only fine tuning on the target domain QA. This indicates knowledge transfer from general domain QA is crucial for technical QA.

\subsubsection{Knowledge transfer across tasks} In Table \ref{tab:ablation}, transferring knowledge across tasks  better capture local correspondence and global semantic relationship between the question and document. Compared with BERT$_{\textbf{RC}}$, TransT improves Ma-F1 by +0.94\% and HA\_F1@1 by +1.91\%.

\subsubsection{Across both tasks and domains} In Table \ref{tab:ablation}, transferring knowledge across both tasks and domains further improve model performance. TransTD fine tunes on SQuAD, then further fine tunes on the TechQA with both RC and AR tasks. It performs better than TransD and TransT. 
TransTD$^{+}$ makes adjustable joint learning, which further brings +1.7\% and +2.32\% improvements on Ma-F1 and HA\_F1@1 compared to TransTD.

\subsubsection{Comparison with retrieve-then-read (two-stage) methods} 
Using semantic similarity to predict alignment between query and document in open-domain QA is an efficient and accurate method. It can be statistical-based (e.g., BM25)~\cite{yang2019end} or neural-based that can be jointly optimized with snippet prediction~\cite{karpukhin2020dense,lee2019latent}. However, as shown in Table \ref{tab:retrieve-then-read}, in the case of the same encoder (i.e., BERT), our proposed TransTD with novel snippet ranking function can identify answers more accurately than above methods. This means that our method is more effective in the context of non-factoid QAs whose semantics of query and document are not aligned.

\subsection{Parameter Analysis}

\paragraph{Loss ratio} In Figure \ref{fig:labelloss}, we compare performance with loss ratio between the RC and DR tasks, $\lambda$ in Eq.(\ref{eq:loss}). We observe that when $\lambda = 4.0$, TransTD achieves the best performance for both RC and DR tasks. If the loss ratio becomes more than $4.0$, the performance decreases significantly. This is because RC helps DR more than DR helps RC, which is consistent with results in Table \ref{tab:ablation}.


\paragraph{Number of fine tuning layers} As shown in Figure \ref{fig:ft_strategy}, we compare performance on different numbers of fine tuning layers. Fine tuning all layers (24 layers) makes the best performance. However, the model performance and the number of fine tuning layers are not an absolute linear relationship. For example, only fine tuning 12 to 14 layers achieves better performance than having 16 or 18 layers, making a good reference for training with limited GPU memories.

\subsection{Error Analysis}

As shown in Figure \ref{fig:error}, we manually categorize the predictive results of 160 answerable question instances in the development set. First of all, there are 107 (64.4\%) questions that can be correctly matched with corresponding documents through the joint inference by Eq.(\ref{eq:joint}), however, 53 (35.6\%) questions are mismatched with the documents that do not contain desirable answers. Additionally, among 107 correct predictions, only 39 (36.4\%) of them are given with the correct answer snippet in the best matching document. Among 68 wrong predictions, 32 (47.1\%) of them are mismatched with the answer span. Besides, 16 (23.5\%) of them are provided with a smaller span of answer snippet than the actual span, in which the average length of answer snippet is 44 words. On the contrary, 20 (29.4\%) of them are provided with a larger span of answer snippet than the actual span, in which their average length is 16 words. We observe that the TechQA dataset offers a challenging yet interesting problem, where the answers have a wide range of the number of words. Some long answers are across multiple sentences.

\section{Conclusion}
\label{sec:conclusion}
In this paper, we studied QA in the technical domain, which was not well investigated. Technical QA faces two unique challenges: (i) the question and answer rarely overlaps substantially (on-factoid questions) and (ii) very limited data size. To address the challenges, we propose a novel framework of deep transfer learning to effectively address TechQA across tasks and domains. To this end, we present an adjustable joint learning approach for document retrieval and reading comprehension tasks. Our experiments on the TechQA dataset demonstrates superior performance compared with non-transfer learning state-of-the-art methods.

\section*{Acknowledgements}
The authors would like to thank the anonymous referees for their
valuable comments and suggestions. This work is supported by National Science Foundation grants, IIS-1849816 and CCF-1901059.


\balance
\bibliography{reference}
\bibliographystyle{acl_natbib}





\end{document}